\documentclass[11pt]{article} 
\usepackage[preprint]{tmlr-style-file-main/tmlr}

\usepackage{amsmath,amsfonts,bm}

\def\eqref#1{equation~\ref{#1}}

\def\1{\bm{1}}

\def\vg{{\bm{g}}}

\def\vx{{\bm{x}}}

\DeclareMathAlphabet{\mathsfit}{\encodingdefault}{\sfdefault}{m}{sl}
\SetMathAlphabet{\mathsfit}{bold}{\encodingdefault}{\sfdefault}{bx}{n}

\DeclareMathOperator*{\argmax}{arg\,max}

\usepackage[ruled, vlined]{algorithm2e}
\usepackage{amsthm}
\usepackage{hyperref}
\usepackage{subcaption}
\usepackage{tikz}
\usepackage{url}

\title{Greedy dynamical meta-learning}

\author{\name Aria Yom \email ariaxiv@proton.me \\
        \addr Rhodora}

\def\month{MM}  
\def\year{YYYY} 
\def\openreview{\url{https://openreview.net/forum?id=XXXX}} 

\SetKwFunction{Sub}{Subroutine}

\begin{document}

\maketitle

\begin{abstract}
    Gradient descent scales well to large models, but becomes unstable over long time horizons. Gradient-free optimizers can scale to arbitrary timespans, but are hobbled by high dimensions. Since learning occurs in large models over long timescales, neither of these approaches is likely to produce traits which can accelerate the learning process. Instead, we propose a meta-learning algorithm in which the agent learns to modify its own weights and biases. Our algorithm consists of an inner loop, wherein the agent performs some high-dimensional optimization upon itself, and an outer loop, wherein we perform some low-dimensional optimization upon the inner loop. Since the outer loop handles very few parameters, standard zeroth-order methods may be used.
\end{abstract}

\section{Introduction}

The child knows little, but can learn with ease. In the ML community, we have long envied this design. Few-shot learning in Large Language Models (LLMs) \citep{Radford_19_Sutskever, Brown_20_Amodei} and the remarkable efficiency of fine-tuning \citep{Donahue_14_Darrell, Yosinski_14_Lipson, Howard_18_Ruder, Devlin_18_Toutanova} are the closest we have come to such adaptability thus far. These capacities scale with network size, and modern neural nets already outperform humans on a variety of tasks \citep{Hestness_17_Diamos, Devlin_18_Toutanova, Kaplan_20_Amodei, Wei_22_Fedus, Schaeffer_23_Koyejo}. Accordingly, the dominant deep learning paradigm today centers around pretraining, fine-tuning, and continual reinforcement learning of large models.

But historically, many have argued that insofar as intelligence is the ability to learn, true intelligence will only be achieved by algorithms that optimize the learning process itself. Such is the meta-learning mantra, although there is some disagreement over what it entails \citep{Thrun_98_Pratt, Li_17_Malik, Hospedales_22_Storkey, Vettoruzzo_24_Thorsteinn}. For our part, we shall focus on approaches which optimize the parameter updates themselves. This includes early works on optimizing synaptic learning rules and self-referential networks \citep{Bengio_95_Cloutier, Runarsson_00_Jonsson, Schmidhuber_93_meta}, but excludes such popular techniques as Siamese Networks and Model-Agnostic Meta Learning (MAML) \citep{Koch_15_Salakhutdinov, Finn_17_Levine}. We recommend the reviews of \citet{Huisman_21_Plaat} and \citet{Hospedales_22_Storkey} and the brief history section of \citet{Andrychowicz_16_De_Freitas} for more background.

Two meta-learning approaches are of particular interest to us: the reinforcement learning (meta-RL) approach \citep{Schmidhuber_99_Wiering, Li_17_Malik} and the self-referential network (SRN) approach \citep{Schmidhuber_92, Schmidhuber_93_meta}. In the meta-RL approach, the learning algorithm is viewed as some parametrized policy $\pi_{\bm{\phi}}$ guiding the evolution of a base-model $\bm{\theta}$. The goal is to devise some \emph{meta}-learning algorithm to adjust $\pi_{\bm{\phi}}$ so as to maximize the growth of a reward function $\mathcal{R}(\bm{\theta})$ over a long time horizon. The hope is that the learned algorithm $\pi_{\bm{\phi}}$ may come to outperform gradient descent in some way.

The problem of course is that this simply replaces the search for $\bm{\theta}$ with the search for $\bm{\phi}$. Should we then pursue a meta-meta-learning algorithm on top of this, and so on? The SRN approach closes this loop by employing a recurrent neural network (RNN) capable of addressing and modifying its own internal parameters. The theory is that the network can embody its own learning algorithm, update its weights as it sees fit, and rewrite its algorithm as it learns new things. Needless to say, this vision has not been realized in experiments, but why?

One reason may be that modern algorithms are simply not suitable for training meta-networks like SRNs. Unlike a typical static model, which stores its knowledge in some fixed weights and biases, the SRN is a dynamical system, storing its knowledge precariously in the orbits of its parameters. As we detail in Secs. \ref{sec:gradients} - \ref{sec:nongradients}, contemporary learning algorithms are not equipped to handle this problem. A new perspective is required, a \emph{dynamic} one.

In Sec. \ref{sec:frame}, we outline a simple framework for the training of dynamical systems. In Secs. \ref{sec:learner} and \ref{sec:cycle}, we develop a sort of evolutionary algorithm for meta-learning in these systems. The main technical difficulty is the handling of dynamical timescales. In Secs. \ref{sec:tuner} and \ref{sec:shape}, we describe an elegant algorithm for tuning these timescales. Readers familiar with the exploding gradient problem and the curse of dimensionality in zeroth-order optimization may skip ahead to these sections, but we suspect that many will find our take on these problems intriguing.

\subsection{Why the gradient won't work for AI} \label{sec:gradients}

Learning is a process which occurs over time. The gradient is an object which compounds over time. In general, if a dynamical system spends $t$ time steps learning something, then the singular values of its derivative will evolve like $\sigma_i \sim e^{\lambda_i t}$. Positive Lyapunov exponents $\lambda_i$ lead to the well-known exploding gradient problem, while negative ones lead to vanishing gradients.

Many of the greatest advancements in deep learning may be viewed as techniques to combat this problem. Rectified linear units (ReLUs), skip connections, and layer normalization are prominent examples \citep{Srivastava_15_Schmidhuber, He_16_Sun, Ba_16_Hinton}. The problem is most pronounced in RNNs, where the gradient must be backpropagated through time \citep{Bengio_94_Frasconi, Pascanu_13_Bengio}. Many workarounds have been proposed, including gradient clipping and unitary matrices for exploding gradients \citep{Pascanu_13_Bengio, Arjovsky_16_Bengio} and non-saturating activations and long short-term memory (LSTM) for vanishing ones \citep{Hochreiter_97_Schmidhuber, Le_15_Hinton, Chandar_19_Bengio}. Ultimately however, the problem can never truly be overcome, as compression and nonlinear transformation of information are the essence of neural computation \citep{Bengio_94_Frasconi}.

It is helpful to view the problem through the lens of chaos theory, wherein the stretching and folding of phase space give rise to the butterfly effect. In chaotic systems, the \emph{predictability horizon} is defined roughly as $T_\lambda = \frac{1}{\max{\lambda_i}}$. Beyond this point, the gradient becomes unstable, and precise forecasting becomes impossible. This is not to say that nothing beyond the horizon can be known, as the probabilities of various outcomes may still be determined through simulation. Rather, the predictability horizon simply represents some limit beyond which the gradient cannot see.

\clearpage

There are essentially two possibilities. If the horizon is distant, and if the learning landscape is barren afar, then there is little need for us to wander beyond the gradient's field of view. But if the horizon is near, and if the world beyond the horizon is rich with possibility, then to rely on the gradient alone as our compass would be to abandon the most fruitful lands to remain forever unexplored.

There are many reasons to believe in the latter possibility. First of all, empirical evidence suggests that the horizon is near. Gradients in RNNs are typically truncated after less than 100 steps, and the active components of residual networks are only tens of layers deep \citep{Veit_16_Belongie, Merity_17_Socher, Zaremba_14_Vinyals}. Second of all, we know that deep networks are capable of things shallow networks are not. It seems unlikely that deep cognition will be possible in less than 100 time steps. Third of all, the best (only?) known examples of superhuman intelligence today come from algorithms that combine deep learning with structured search \citep{Silver_17_Hassabis, Schrittwieser_20_Silver, Hafner_20_Ba, Hubert_26_Silver}. The search, it would appear, covers a key blind spot of the gradient, but it is not clear that this is the only blind spot, and it is unlikely that these techniques can be generalized.

All this leads us to a simple conclusion, that gradient descent is unfit to produce learning systems. From this perspective, models today should be regarded not as \emph{intelligent}, but merely as \emph{performant}. To further clarify this point, and to foreshadow our solution, let us consider a simple example:

\subsection{Example: the tortoise, the hare, and the taffy}

In this section we show that simple sampling beats gradient descent in chaotic systems. Consider the following variant of the doubling map over $x_n \in (-1,1)$, with $p,q \in (0,1)$ (Fig. \ref{fig:taffy}):

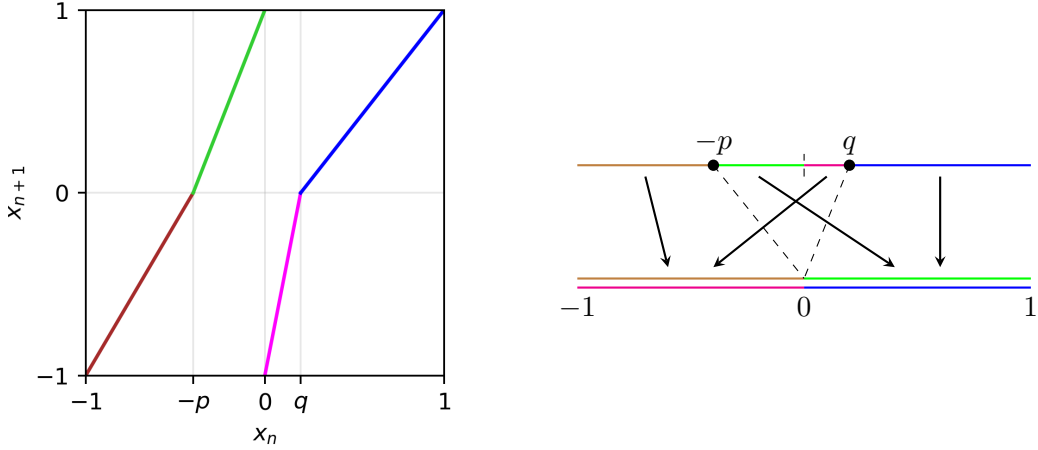
\begin{figure}
    \centering
    
    \begin{subfigure}[c]{0.4\textwidth}
        \centering
        \includegraphics[width=\linewidth]{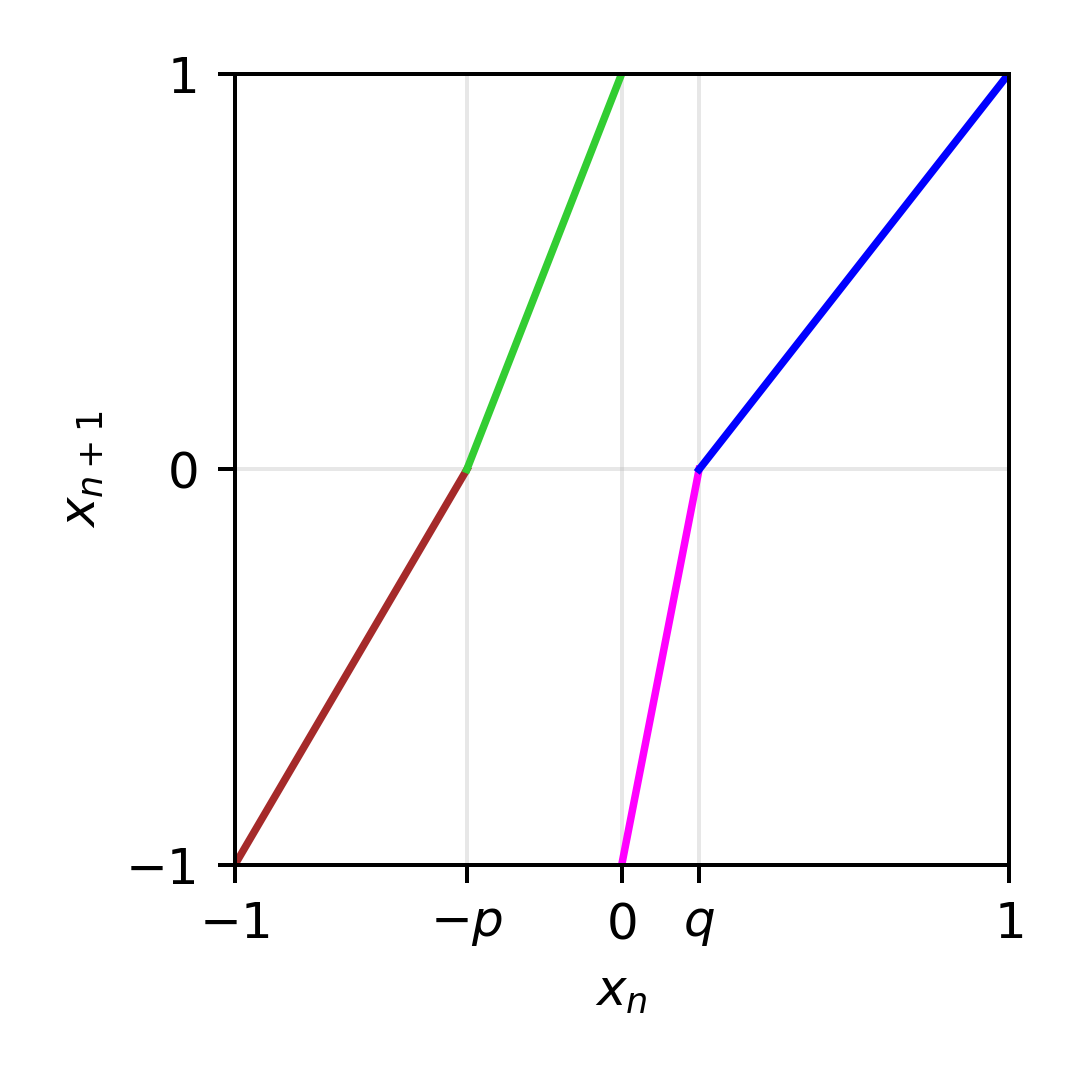}
    \end{subfigure}
    \begin{subfigure}[c]{0.5\textwidth}
        \centering
        \begin{tikzpicture}[scale=3, >=stealth]

            \def\p{0.4}
            \def\q{0.2}
            \def\z{-.5}

            \draw[brown, thick] (-1,0) -- (-\p,0);
            \draw[green, thick] (-\p,0) -- (0,0);
            \draw[magenta, thick] (0,0) -- (\q,0);
            \draw[blue, thick] (\q,0) -- (1,0);

            \draw[brown, thick] (-1, \z) -- (0, \z);
            \draw[green, thick] (0, \z) -- (1, \z);
            \draw[magenta, thick] (-1, \z - .04) -- (0, \z - .04);
            \draw[blue, thick] (0, \z - .04) -- (1, \z - .04);

            \fill (-\p,0) circle (0.025);
            \fill (\q,0) circle (0.025);

            \node[above] at (-\p, 0) {$-p$};
            \node[above] at (\q, 0) {$q$};
            \node[below] at (-1, \z - .04) {$-1$};
            \node[below] at (0, \z - .04) {$0$};
            \node[below] at (1, \z - .04) {$1$};

            \draw[dashed] (-\p,0) -- (0,\z);
            \draw[dashed] (\q,0) -- (0,\z);
            \draw[dashed] (0,-.05) -- (0,.05);

            \draw[->, thick] (-\p/2, -0.05) -- (.4, \z + 0.05);
            \draw[->, thick] (\q/2, -0.05) -- (-.4, \z + 0.05);
            \draw[->, thick] (-1/2 - \p/2, -0.05) -- (-0.6, \z + 0.05);
            \draw[->, thick] (1/2 + \q/2, -0.05) -- (0.6, \z + 0.05);

        \end{tikzpicture}
    \end{subfigure}
    
    \caption{The taffy map.}
    \label{fig:taffy}
\end{figure}

\begin{align}
    x_{n+1} = T_{p,q}(x_n) = 
    \begin{cases} 
        \frac{x_n + p}{1 - p} & \quad -1 < x_n < -p \\
        \frac{x_n + p}{p} & \quad -p < x_n < 0 \\ 
        \frac{x_n - q}{q} & \quad  0 < x_n < q \\
        \frac{x_n - q}{1 - q} & \quad q < x_n < 1 
    \end{cases}
\end{align}

This map is equivalent to cutting a piece of taffy in half, stretching the two halves alongside each other, and remixing them together (Fig. \ref{fig:taffy}). After many cycles, the probability density of a point winding up in either half of the taff will be:

\begin{align}
    \rho(x) \approx \label{down}
    \begin{cases}
        \frac{q}{p + q} & \quad -1 < x < 0 \\ 
        \frac{p}{p + q} & \quad 0 < x < 1 
    \end{cases}
\end{align}

Let us consider the simple reward function $\mathcal{R}(x) = x$. We would like to adjust $p$ and $q$ so as to maximize the long-term reward $R = \lim_{n \rightarrow \infty} \frac{1}{n} \sum_{i = 1}^n \mathcal{R}(x_i)$. The gradient of $R$ may be estimated in two ways, either by backpropagation or by a zeroth-order approximation:

\begin{align}
\bm{g}_1(p, q, N) &= \mathbb{E} \left[ \frac{1}{N} \sum_{i = 1}^N \nabla_{p,q} \mathcal{R}\big(T^i_{p,q}(x_0)\big) \right] \\
\bm{g}_0(p, q, N) &= \mathbb{E} \left[ \frac{1}{N} \sum_{i = 1}^N  \frac{\mathcal{R}\big(T^i_{p+ \delta p ,\, q + \delta q}(x_0)\big) - \mathcal{R}\big(T^i_{p- \delta p ,\, q - \delta q}(x_0)\big)}{|\delta \bm{\theta}|^2} \delta \bm{\theta}\right]
\end{align}

where the expectation is taken over $x_0$ and $\delta \bm{\theta} = (\delta p, \delta q)$, and where $\delta \bm{\theta}$ is drawn uniformly from a sufficiently small circle.

\begin{figure}
     \centering
     \hspace{-1cm}
     \includegraphics[width = \columnwidth]{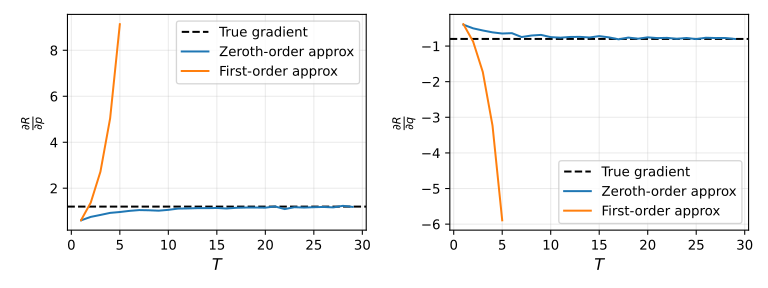}
     \caption{First-order and zeroth-order gradient estimates at $(p ,\, q) = (.2 ,\, .3)$ for $N = 10^4$. $x_0$ was sampled from the invariant measure (Eqn. \ref{down}), while $(\delta p, \delta q)$ was drawn from a circle of radius $.1$.}
     \label{fig:compare}
\end{figure}

But we know from Eqn. \ref{down} that the reward should be $R = \frac{p-q}{2 p + 2 q}$, so it is easy to compare these two approaches to the analytic solution and see which is more accurate. From Fig. \ref{fig:compare} we can see that the zeroth-order estimate slowly and steadily converges, while the first-order estimate sprints off to infinity and gets lost in the taffy.

The source of this behavior is the hidden randomness of the map, which the gradient cannot see, but which the statistics always reveal. Our claim is that many cognitive processes, such as search, have an essential randomness to them which only trial and error can capture. If our learning agents were as two-dimensional as this example, then there would be little more to say, but as is well-known, zeroth-order approaches face quite a challenge in higher dimensions:

\subsection{The curse of dimensionality} \label{sec:nongradients}

Nongradient methods appropriate for many-parameter stochastic optimization all share a basic structure. They consist of some scheme for randomly sampling parameters $\bm{\theta}$ along with some rule for iteratively improving these samples over time. The problem of course is that when parameter space is large, there are always many more ways to decay into a high-entropy state than to improve into a high-performance state.

To see this, suppose that the loss near $\bm{\theta}$ is locally quadratic, such that small perturbations $\delta \bm{\theta}$ change the loss by roughly:

\[
    \delta\mathcal{L} \approx \vg \cdot \delta \bm{\theta} + h |\delta \bm{\theta}|^2
\]

If $\delta \bm{\theta}$ points in a random direction in $\mathbb{R}^d$, the gradient term will contribute roughly $\pm \frac{|\vg|} {\sqrt{d}}|\delta \bm{\theta}|$, while the Hessian term will incur a cost of $h|\delta \bm{\theta}|^2$. Thus, large steps will always result in a loss of performance, and the optimal step size will be roughly $|\delta \bm{\theta}| \sim \frac{|\vg|}{h\sqrt{d}}$. The improvement per step will go like $\delta \mathcal{L} \sim -\frac{1}{d}$.

The situation only worsens with the introduction of noise. If the loss measurement is noise dominated, which it inevitably will be at these small scales, then the number of samples required to resolve a difference will scale inversely with $\delta \mathcal{L}^2$. Thus, it will take $O(d^2)$ measurements to estimate the gradient. If the noise scales with $d$, then the penalty could be even worse. 

Evidently, for a billion-parameter model, this price is unpayable. For this reason, many algorithms have been devised to reduce the dimensionality of the search space \citep{Hansen_95_Gawelczyk, Ros_08_Hansen, Wang_16_Freitas, Maheswaranathan_19_Sohl_Dickstein, Back_23_Ye}. One example is the Covariance Matrix Adaptation Evolution Strategy (CMA-ES), which shrinks mutations along axes with high loss curvature and grows mutations in directions that produce beneficial adaptations \citep{Hansen_96_Ostermeier, Hansen_01_Ostermeier}. In so doing, it attempts to squeeze mutations orthogonal to the sweep of the gradient out of existence, leaving only the components in a critical low-dimensional subspace. The main problems with this type of approach are that (a) the size of the covariance matrix scales like $O(d^2)$, (b) the mutations must be normally distributed, and (c) the learning landscape must be relatively stable. We too would like to optimize the shape of the mutation distribution, but in order to overcome these challenges, we take a different approach.

The problem of finding a good mutation distribution is no different from any other problem we face in machine learning. We are seeking some low-dimensional manifold in a high-dimensional space. We cannot write down an equation for it. Strategies for where and how to search and tactics for compressing the space amount to little more than feature engineering. They may see some initial success, but are ultimately doomed \citep{Sutton_19}. The only path forward is to train a system to discover the optimal mutation distribution for us. The only path forward is meta-learning.

\clearpage

\subsection{Dynamical systems meta-learning} \label{sec:frame}

We are now finally in a position to explain our meta-learning algorithm. Let us begin with an overview of the dynamical systems meta-learning (DSML) framework. A dynamical system is a state $\vx(t)$ governed by some fixed laws of motion, such as Hooke's law $\vx'' = -\vx$. In our case, these dynamics must be stochastic, so as to generate discernable variations in $\vx(t)$. It is important to understand that both the weights and activations are contained within $\vx$ and will evolve over time. A DSML protocol consists of three parts:

\begin{enumerate}
\item The dynamics $\vx' = \mathcal{D}(\vx)$
\item The reward function $\mathcal{R}(\vx)$
\item The training algorithm $\mathcal{A}(\vx, \mathcal{R}, \mathcal{D})$
\end{enumerate}

In the interest of brevity, in this paper we will focus only on the training algorithm. We will have more to say about the system architecture and dynamical equations in an upcoming publication, and we have quite little to say about how to devise a good reward function.

\begin{figure}
     \centering
     \includegraphics[width = \columnwidth]{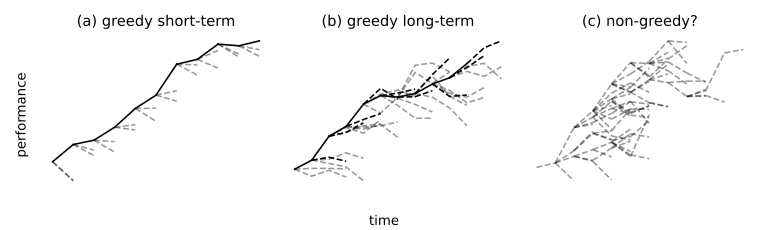}
     \caption{Time and greed. In the short-term case, $3$ mutants are generated at each time step, with the best surviving and reproducing. The long-term plot is similar, but the system is simulated $3$ steps into the future before crowning the winner. In a more realistic scenario, ``long-term'' would likely mean much more than $3$ steps. Gray dashed lines denote paths sampled. Black dashed lines denote the winning paths from each branch point. The solid line denotes the optimization trajectory. Note that the winning paths extend beyond the solid line.}
     \label{fig:greed}
\end{figure}

Our approach will be to simulate the system along many trajectories and select for the most favorable ones. The simplest algorithm of this form is probably the $(1, \lambda)$-ES algorithm, which produces a clutch of $\lambda$ mutants in each generation and selects the top performer to sire the next generation \citep{Beyer_02_Schwefel}. This algorithm is illustrated in Fig. \ref{fig:greed}a.

We call such an approach \emph{greedy}, as it always exploits the best mutant and never explores other branches. It is also \emph{short-term} in the sense that it evaluates each mutant at its current fitness rather than considering its potential growth over a long time horizon. The $(1, \lambda)$-ES algorithm can easily be modified into a longer-term form by simulating each trajectory out several steps into the future and evaluating each mutant by its late-time fitness. This is illustrated in Fig. \ref{fig:greed}b, and is essentially the algorithm we will be developing in Secs. \ref{sec:learner} and \ref{sec:cycle}.

It is worth noting that both of these greedy approaches may be vulnerable to getting stuck in local optima. Ideally, we would employ a non-greedy algorithm to perform some judicious exploratory rollouts (Fig. \ref{fig:greed}c). Unfortunately, we were not able to devise a principled algorithm along these lines, but the lesson of deep learning seems to be that large networks are fairly unencumbered by local optima, so hopefully this will hold for DSML as well.

The difference between the DSML and meta-RL frameworks is that in meta-RL the time evolution is determined by a trainable policy, whereas in DSML, it is given by some immutable laws of motion. Our agents have no agency to rewrite these laws. Furthermore, their performance will likely depend on these laws being well-designed. So it may appear that we are simply inventing a worse version of meta-RL, one devoid of any ability to improve its policy, one with even more hand-crafted features, and one inherently vulnerable to local optima. How on Earth is this supposed to be better?

The lack of a trainable policy is not a bug, it is a feature. In meta-RL, the system parameters are divided between the model and the policy. Nothing the model learns can be used to refine its evolution. By contrast, our goal is to create an agent that gains knowledge from the training data and utilizes that knowledge to improve its own mutations. If it can squeeze these mutations down to a few key axes, then it will be no trouble for the evolutionary razor to distinguish the best mutants.

By eliminating the policy parameters, we force all of the meta-information into the system state $\vx$ itself. When trained over a long time horizon, it becomes necessary for the agent not only to perform well, but to produce mutations which enhance its performance over the measured horizon, or at least to avoid mutations which may destroy its knowledge. But how long should the time horizon be? What follows is essentially a study in the handling of this and other timescales.

\subsection{Summary}

The preceding argument may be summarized as follows:

\begin{enumerate}
\item Learning is a process which occurs over long timespans.
\item The gradient becomes unstable over long timespans. It is magical thinking to believe that the butterfly effect can be overcome by some clairvoyant mathematics.
\item The only way to learn how to learn is to generate mutations at random, evaluate them over long timescales, and select the best learners.
\item Unfortunately, the concept of randomness presupposes an underlying distribution, and it is not clear what the mutation distribution should be. Any attempt to hand-craft a clever mutation distribution is ill-fated.
\item We would instead like the agent itself to learn how to generate its own mutations.
\item Our solution is to encode the agent in a stochastic dynamical system. The system state then determines how its weights will evolve, subject to some laws of motion.
\item By selecting high-performing agents over long timescales, we will also be selecting for those which routinely generate beneficial variations, preserve learned structures, and recover from damaging mutations.
\end{enumerate}

\clearpage

\section{Results} \label{sec:results}

\subsection{Mutation and evaluation timescales} \label{sec:learner}

Suppose our system begins in some low-intelligence, high-entropy attractor. In order to learn, it will need to escape this attractor. Since the dynamics are stochastic, there will always be some trajectories which randomly jump some distance out of the heart of the attractor before sliding back in. Our goal is to find these trajectories, truncate them at their apexes before falling back down, and respawn new trajectories from these higher intelligence states. By iterating this process, we can slowly ascend up the walls of the attractor basin, potentially crossing into the basins of more intelligent attractors along the way.

Unfortunately, this is only possible if we can measure the intelligence, which we have no direct means of doing. The hope is that it may suffice to find a good reward function $f$. If we identify the intelligence roughly with the growth rate of the reward, then $f$ will be a \emph{lagging} measure of the intelligence. Thus, the distinction between the latent intelligence and the measured performance is not merely philosophical. There are two distinct timescales: the optimal mutation time $\mu$, at which the apexes of the intelligence occur, and the optimal evaluation time $\nu$, at which the performances are maximally distinguishable.

In order for learning to occur, these two timescales must be properly calibrated. If $\mu$ or $\nu$ are too small, then the trajectories will not have had enough time to meaningfully diverge. But if they are too large, then the trajectories will fall back down into the nearest attractor. In the case of $\mu$, this will result in a loss of intelligence, whereas in the case of $\nu$, this will result in the trajectories becoming indistinguishable. Thus, the intelligence exists and is observable neither in the short term nor in the long term, but in the medium term.

In order to better understand these points, let us consider a simple toy example. Suppose the ``intelligence'' $z$ evolves according to the Ornstein-Uhlenbeck equation $dz = -zdt + dW$. In the absence of any selection pressure, $z$ will always come to hover around the attractor at $z = 0$. Now suppose the measurable performance $y$ of the system evolves according to $\frac{dy}{dt} = -y + z^3$, such that $z$ influences how quickly the performance grows \footnote{The simpler $\frac{dy}{dt} = -y + z$ produces essentially the same results, but the cube helps visually exaggerate the effect.}. What do the dynamics of such a system look like?

\begin{figure}
     \centering
     \includegraphics[width = \columnwidth]{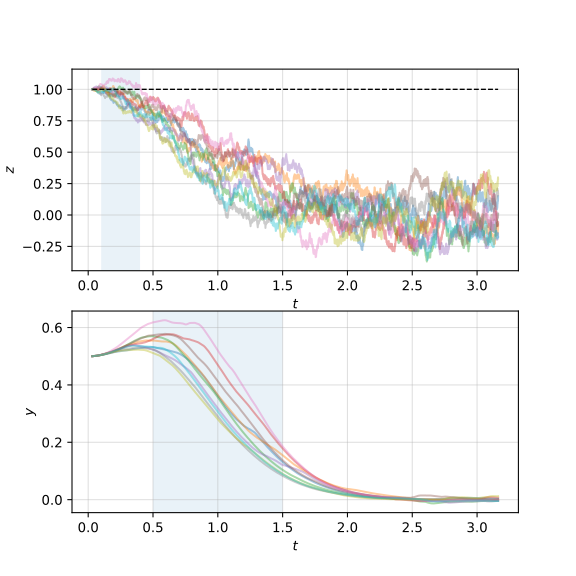}
     \caption{Simulation of the system $dz = -z dt + dW$, $dy = (-y  + z^3) dt$, for a Wiener process $W$. Initial conditions $z = 1$ and $y = .5$ were used for $10$ random trajectories. The optimal mutation timescale $\mu$ is highlighted in the upper $z$ plot, while the optimal evaluation timescale $\nu$ is highlighted in the lower $y$ plot.}
     \label{fig:sketch}
\end{figure}

In Fig. \ref{fig:sketch}, we plot several random trajectories of this system diverging from the point $ (t, z, y) = (0, 1, \frac{1}{2})$. We choose the high starting intelligence $z = 1$ so as to emulate being in the middle of a training run. Evidently, the pink trajectory is best, with maximal intelligence occurring around $\mu \approx .25$. But the performance is maximally distinguishable around $\nu \approx 1$. So if we had chosen to produce the next generation of mutants from the state $x_\text{pink}(t = \nu)$, we would have realized a loss of intelligence. In fact, if we had chosen the absolute maximal performance state $x_\text{pink}(t =.6)$, we still would have realized a loss of intelligence.

In other words, by the time the trajectories can be distinguished, the best-performing trajectory may already have lost the qualities that enabled it to gain such distinction. We cannot know a priori how far $\nu$ will lag behind $\mu$. This gap is DSML's raison d'etre. It is neither addressed by static models, which fail to evolve over the course of their evaluations, nor by evolutionary algorithms, which spawn new mutants directly from the evaluated state.

\subsection{Learning cycle} \label{sec:cycle}

With the two timescales $\mu$ and $\nu$ in mind, we may construct the following simple evolutionary algorithm: Every generation begins with a state $x_\text{parent}$, from which $m$ mutants $x_i(t)$ are spawned, with $x_i(0) = x_\text{parent}$. From each mutant, $n$ evaluation trajectories $y_i^j(t)$ are spawned, with $y_i^j(\mu) = x_i(\mu)$. These trajectories are then evaluated $k$ times at the optimal timescale, $r_{i,1}^j \dots r_{i,k}^j \overset{\text{iid}}{\sim} f(y_i^j(\nu))$, and the mutant with the greatest mean performance $z_i = \frac{1}{nk} \sum_{j,l} r_{i,l}^j$ is chosen to sire the next generation.

Much like $\mu$ and $\nu$, the parameters $m, n, k$ must be properly balanced in order to ensure the algorithm runs efficiently. Taking too many samples is wasteful, while taking too few will lead to degenerating performance. Thus, there are five parameters we must balance in our algorithm:

\begin{description}
\item[$\mu$ = ] How long is a generation?
\item[$m$ = ] How many mutants should we spawn per generation?
\item[$n$ = ] How many trajectories should we sample per mutant?
\item[$k$ = ] How many evaluations should we perform per trajectory?  
\item[$\nu$ = ] How far into the future should we perform the evaluations?
\end{description}

Our solution to the first of these Goldilocks problems can be found in the next section. The second is given by \citet{Rechenberg_73}. In this section, we assume that $\mu$ and $m$ are fixed, and show that the last three parameters $n, k, \nu$ can all be combined into one using the classic $\int \frac{1}{t} = \log t$ trick.

To begin, let us note that the intent of the sample mean $z_i = \frac{1}{nk} \sum_{j,l} r_{i,l}^j$ is to estimate the true mean $\mathbb{E}[f(y_i(\nu))]$ of a trajectory spawned from $y_i(\mu) = x_i(\mu)$. Measuring a single trajectory many times reduces the measurement variance, but does nothing to address the intertrajectory variance. Measuring many trajectories once each produces a better estimate of the mean, but if the cost of simulating each trajectory out to $t = \nu$ is large, then this approach is very wasteful. The optimal parameters $n$ and $k$ arise from trying to obtain the best estimate of the mean $\mathbb{E}[f(y_i(\nu))]$ for the least computational cost.

Suppose that we have found the optimal $n, k, \nu$, and let us assume that simulating a single timestep costs the same amount of compute as performing a single $f$ evaluation. Then our optimized algorithm will cost $n(k + \nu - \mu)$ units of compute per mutant. We will now show that we can avoid the matter of finding $n, k, \nu$ altogether with a sampling scheme which is only logarithmically worse than the optimum.

Suppose we simulate each trajectory $y_i^j(t)$ out to $t = \frac{\tau}{j}$ for each $j = 1 \dots \tau$ \footnote{We encourage the reader not to think too much about whether these times are discrete or continuous.}. Then, for each trajectory, we perform an $f$ evaluation at every timestep $t > \mu$ (we may discard the trajectories which terminate before $t = \mu$). The amounts of compute spent on simulations and evaluations in this scheme will be roughly the same, and neither will exceed $\tau \log \tau$.

The key point of this approach is that the amount of compute it spends at every timescale is the same \footnote{It is important to note that when we say ``timescale'', we really mean something like $\log(t) \pm 1$, rather than $t \pm 1$.}. In particular, over the optimal timescale $t \in \left( \frac{1}{2} \nu, 2 \nu \right)$, it samples roughly $\frac{\tau}{\nu}$ trajectories $\nu$ times each. Thus, if we simply take $\tau = \max \{ n k, n \nu \}$, then our logarithmic algorithm will take at least as many samples around $\nu$ as the optimal scheme, and it will spread them out over at least as many trajectories. But, the excess cost from sampling over all timescales is only a factor of $\log \tau$ \footnote{Although there is potentially an additional log associated with the excess noise coming from these other timescales.}. This approach is summarized in Alg. \ref{alg:trunx}.

\begin{algorithm}
    \DontPrintSemicolon
    \caption{Learning cycle $C_{\mu \tau}$} \label{alg:trunx}
    \BlankLine
    \KwIn{parent $x_\text{parent}$, timescales $\mu, \tau$, stochastic evaluator $f(x)$, stochastic mutator $g(x)$}

    \BlankLine

    $m = 8$ \tcp{Rechenberg's rule} 
    $x_1 \dots x_m \overset{\text{iid}}{\sim} g^\mu(x_\text{parent})$ \tcp{sample mutants}

    \BlankLine
    \tcp{run tests}
    \BlankLine
    \For{$i = 1 \dots m$} {
        $z_i \gets 0$ \;
        \For{$ j = 1 \dots \tau $} {
            $y_i^j \gets x_i$ \;
            \For{$ t = \mu \dots \frac{\tau}{j} $} {
                $y_i^j \gets g(y_i^j)$ \;
                $z_i \gets z_i + f(y_i^j)$ \;
            }
        }
    }
    \BlankLine
    \tcp{crown winner}
    \BlankLine
    $i_\text{max} \gets \argmax_i z_i$ \;
    $x_\text{parent} \gets x_{i_\text{max}}$ \;
    
    \BlankLine
    \KwOut{$x_\text{parent}$}
    \BlankLine
\end{algorithm}

The algorithm should work under two conditions: First, it must not be necessary to evaluate the trajectories at the exact time $\nu$. Measurements at the same time\emph{scale} around $t \in \left( \frac{1}{2} \nu, 2 \nu \right)$ must also suffice to distinguish the trajectories, which seems only logical (Fig. \ref{fig:sketch}). Second, the measurement error must not vary too much with the timescale. If it diverges as $t \rightarrow \infty$ or as $t \rightarrow 0$, then although we may take sufficiently many samples around $t = \nu$, our signal will be lost in the noise from other timescales. This condition should be satisfied by any well-behaved reward function.

So we have managed to combine the parameters $n, k, \nu$ into a single timescale $\tau$, paying only polylogarithmically more than if we had used the optimal parameters. Furthermore, we know from Rechenberg's rule that for well-behaved mutations, we may generally take $m \approx 10$ \citep{Rechenberg_73, Beyer_02_Schwefel}. Thus, the only parameters which remain to be optimized are the two timescales $\mu$ and $\tau$. 

\subsection{Tuning algorithm} \label{sec:tuner}

In the previous section, we derived a learning subroutine $x \rightarrow C_{\mu \tau}(x)$ to iteratively improve the performance of a dynamical system with respect to an objective $f(x)$. The parameters $\mu^{-1}$ and $\tau$ each quantify a sort of evolutionary selection pressure and a corresponding computational cost. In general, we imagine that under sufficient pressure, the system will evolve towards a high-performance state, but it may take a very long time to do so. The challenge therefore is to adjust $\mu$ and $\tau$ so as to maximize performance whilst minimizing computational cost.

Our parameter tuning algorithm consists of running a series of learning cycles $x_{i+1} = C^{\circ T}_{\mu \tau}(x_i)$ in which we apply our subroutine $T$ times and subsequently evaluate our system over $N$ samples. We then use these samples to readjust the parameters $\mu$ and $\tau$ so as to maximize the quantity $ Q(\mu, \tau) = \frac{\mathbb{E}[f(C_{\mu \tau}(x)) - f(x)]}{\tau} $, which corresponds roughly to the training speed. This can be done by solving for $dQ = 0$:

\begin{gather}
dQ = d\left(\frac{ \mu S}{\tau}\right) = \frac{\mu}{ \tau } \left( dS + \left(\frac{d \mu}{\mu} - \frac{d \tau}{\tau} \right) S \right) = 0 \\
dS = S (d \log \tau - d \log \mu) \label{best}
\end{gather}

where we have defined $S(\mu, \tau) = \frac{\mathbb{E}[f(C_{\mu \tau}(x)) - f(x)]}{\mu}$ for reasons which will become clear in the next section.

The right hand side of Eqn. \ref{best} is easily estimated from samples of $S$, while the gradient on the left must be estimated from a zeroth-order method. We will be using random direction stochastic approximation (RDSA), but since there are only two parameters, essentially any other standard optimizer would suffice \citep{Ermoliev_69, Kushner_78_Clark}.

RDSA involves running two cycles in parallel with slightly different parameters and trying to measure the difference in $S$ between them. Suppose we draw $\xi^\mu_i, \xi^\tau_i \overset{\text{iid}}{\sim} \mathcal{N}(0, \sigma^2)$, and let $\mu_i^\pm = \mu_i \exp( \pm \xi^\mu_i)$ and $\tau_i^\pm = \tau_i \exp( \pm \xi^\tau_i)$. Each cycle produces an endstate $x_i^\pm = C^{\circ T}_{\mu_i^\pm, \tau_i^\pm} (x_i)$ with sample mean rewards $z_i^\pm = \frac{1}{N} \sum_{j = 1}^N f(x_i^\pm)$ which may be compared to the original $z_i = \frac{1}{N} \sum_{j = 1}^N f(x_i)$. We then update the parameters as follows:

\begin{align}
    \log \mu_{i+1} &= \log \mu_i +  \eta\left(\frac{1}{\sigma^2}(S^+_i - S^-_i) \xi^\mu_i + (S^+_i + S_i^-) \right) \label{mu_up} \\
    \log \tau_{i+1} &= \log \tau_i +  \eta\left(\frac{1}{\sigma^2}(S^+_i - S^-_i) \xi^\tau_i - (S^+_i + S_i^-) \right) \label{tau_up} \\
    S_i^\pm &= \frac{z_i^\pm - z_i}{T \mu_i \exp(\pm \xi^\mu_i)}
\end{align}

Eqns. \ref{mu_up} and \ref{tau_up} are the heart of our adaptive tuning algorithm (Alg. \ref{alg:main}). $x_{i+1}$ may be chosen randomly from $x_i^+$ and $x_i^-$, although choosing the one with the higher reward may be slightly better. All that is left is to make some good choices for the hyperparameters $\eta, \sigma^2, T, N$.

\begin{algorithm}
    \DontPrintSemicolon
    \caption{Parameter tuning} \label{alg:main}
    \BlankLine
    \KwIn{initial state $x$, stochastic evaluator $f(x)$, learning cycle $C_{\mu \tau}(x)$, learning rate $\eta$}
    \BlankLine

    $\mu, \tau \gets \dots$ \tcp{initializations}
    $\sigma^2 = .1$ \;

    \BlankLine
    \For{$i = 1,2,3 \dots$} {
        $ T \gets \frac{\tau}{\mu} $\;
        $N \gets m \tau T$ \;

        \BlankLine
        \tcp{learning cycles}
        \BlankLine
        $\xi^\mu, \xi^\tau \overset{\text{iid}}{\sim} \mathcal{N}(0, \sigma^2)$ \;
        $\mu^\pm \gets \mu \exp( \pm \xi^\mu )$ \;
        $\tau^\pm \gets \tau \exp( \pm \xi^\tau )$ \;
        $x^\pm \gets C^{\circ T}_{\mu^\pm \tau^\pm}(x) $ \tcp{T loops of subroutine}

        \BlankLine
        \tcp{evaluations}
        \BlankLine
        $y_1 \dots y_N \overset{\text{iid}}{\sim} f(x)$ \;
        $z \gets \frac{1}{N} \sum_j y_j$ \;
        $y_1^\pm \dots y_N^\pm \overset{\text{iid}}{\sim} f(x^\pm)$ \;
        $z^\pm \gets \frac{1}{N} \sum_j y_j^\pm$ \;        
	
        \BlankLine
        \tcp{updates}
        \BlankLine
        $x \gets x^+ \text{ or } x^-$ \;
        $S^\pm = \frac{z^\pm - z}{T \mu^\pm}$ \;
        $\mu \gets \mu \exp \left[ \eta\left(\frac{1}{\sigma^2}(S^+ - S^-) \xi^\mu + S^+ + S^- \right) \right]$ \;
        $\tau \gets \tau \exp \left[ \eta\left(\frac{1}{\sigma^2}(S^+ - S^-) \xi^\tau - S^+ - S^- \right) \right]$ \;
    }
    \BlankLine
    \KwOut{$x$}
    \BlankLine
\end{algorithm}

Let us begin with $T$ and $N$. Since the shape of the learning landscape will inevitably change during training, we would like to update our $\mu$ and $\tau$ as frequently as possible. However, since our system is always being evaluated some $\tau$ steps into the future, we should probably wait at least that long for the effects of any changes in $\mu$ and $\tau$ to manifest before reevaluating. Thus $T = \frac{\tau}{\mu}$ seems like a reasonable choice. For $N$, bigger is always better, but we would like to spend the bulk of our compute on learning cycles rather than parameter tuning, and each cycle contains roughly $m \tau \log \tau$ evaluations, so $N \sim m \tau T$ seems reasonable.

This leaves only two remaining hyperparameters, and as we shall see in the next section, $\sigma^2$ can generally be taken to be around $\sigma^2 \approx .1$. Only $\eta$ is likely to require any difficult tuning. Larger values of $\eta$ adapt more quickly to changes in the optimal $\mu$ and $\tau$. Smaller values of $\eta$ converge more tightly on stationary optima. Ideally, if the learning landscape does not change too rapidly, a wide range of $\eta$ should satisfy both of these demands.

\subsection{The shape of $Q$} \label{sec:shape}

In constructing Alg. \ref{alg:main}, we made a number of unexplained design choices. Why did we pull a factor of $\mu$ into $S = \frac{\mathbb{E}[f(C_{\mu \tau}(x)) - f(x)]}{\mu}$? Why the logarithms $\log \mu$ and $\log \tau$? Doesn't $\sigma^2$ need to be tuned as well? And can the optimal timescales $\mu$ and $\tau$ even be found without getting stuck in local optima? In order to see the answers to these questions, we will need to obtain more insight into the relationship between $\mu$, $\tau$, and $S$.

It should not be difficult to believe that $S$ will increase monotonically with either $\theta \in \{\mu^{-1}, \tau\}$. After all, each $\theta$ represents a sort of multiplicative computational expense that contributes to the selection pressure on the population. $\mu^{-1}$ represents the number of generations per unit time, and $\tau$ represents the number of evaluations per mutant. Presumably, the higher the selection pressure, the sharper the results. We will further argue that $S$ should typically be concave in $\sqrt{\theta}$. It is of course easy to provide counterexamples to this claim \footnote{In fact one can provide counterexamples to the monotonicity and saturation as well...}, but let us consider the context in which these parameters act:

We begin with $\mu^{-1}$. Suppose that a mutant's performance follows a decreasing random walk $\Delta f \overset{\text{iid}}{\sim} \mathcal{N}(-\alpha \Delta t, \beta^2 \Delta t)$ over the mutation time $\mu$. A mutant one sigma above the mean will have $ \Delta f \approx -\alpha \mu + \sqrt{\beta^2 \mu}$, and a corresponding $S \approx -\alpha + \sqrt{\frac{\beta^2}{\mu}}$. So long as we can find such a mutant, shrinking $\mu$ yields gains of $\mu^{-\frac{1}{2}}$. But eventually differences in $\Delta f$ become too small to distinguish, so $S$ saturates to some constant value.

Now consider $\tau$. $S$ is estimated from the mean of roughly $\tau$ samples, and will thus have some error proportional to $\tau^{-\frac{1}{2}}$. When distinguishing between two mutants, the probability of selecting the best one will initially grow like $\frac{1}{2} + \sqrt{\tau}$ before saturating to $1$. This is exactly the same behavior as in $\mu^{-1}$, with $S$ initially growing proportionally to the square root, and eventually levelling off.

\begin{figure}
     \centering
     \includegraphics[width = \columnwidth]{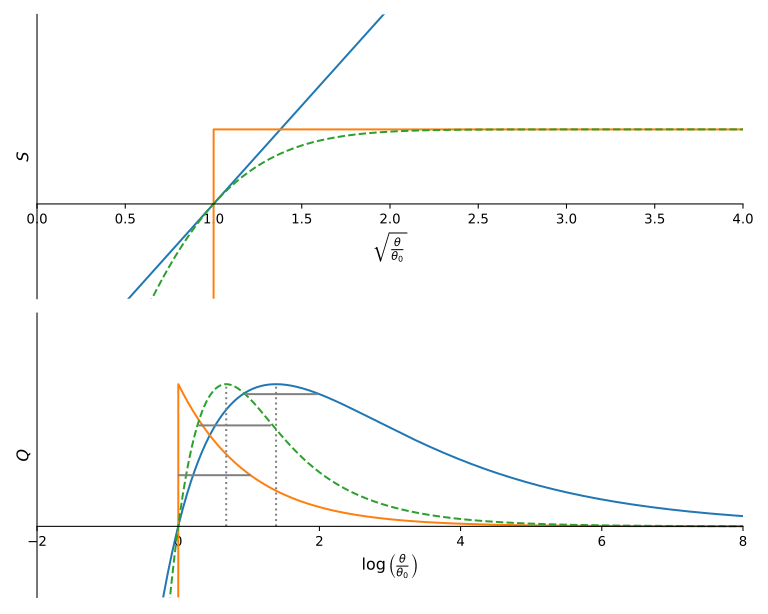}
     \caption{Shapes of $S$ and $Q$. The solid blue line represents the limiting case where $S \propto \sqrt{\theta} - \sqrt{\theta_0}$. The solid orange line represents the limiting case where $S$ saturates to its maximum immediately upon crossing $\theta > \theta_0$. The dashed line represents a ``typical'' case in between these two limits. The gray horizontal lines are of unit length and show that fluctuations of size $\sigma \ll 1$ should be well-tolerated for any such concave $S$.}
     \label{fig:sqrt}
\end{figure}

The implication of this is that along any parameter axis one may presume that $S$ sits between two limiting concave forms, one in which $S \sim \sqrt{\theta} - \sqrt{\theta_0}$ saturates far away from its zero $\theta_\text{sat} \gg \theta_0$, and one in which $S$ acts like a step function at $\theta_0$, saturating immediately at $\theta_\text{sat} \approx \theta_0$. These two limits are illustrated in Fig. \ref{fig:sqrt}. Crucially, in both cases, when plotted against $\log \theta$, the \emph{shapes} of these curves, and in particular their widths, are unaffected by changes in $\theta_0$. Furthermore, the concavity implies that $Q$ has a unique local maximum.

In studying these figures one is struck by the realization that a single $\sigma^2$ may be prescribed for any $Q$ satisfying these assumptions! From Fig. \ref{fig:sqrt}, we can see that fluctuations in $\log \theta$ of about $.5$ around the optimum will reduce performance by less than a factor of 2. So long as $\sigma$ is reasonably smaller than this, then in the steady state our tuner should track quite close to the optimum. Of course, the larger we can make $\sigma$, the better our resolution of the gradient will be, and the faster our tuner will adapt to changes. Thus, we may conclude that $\sigma^2 \approx .1$ should be a reasonable choice for any well-behaved learning landscape.

It is important to note that the slope $\frac{dQ}{d \log\theta}$ goes to zero as $\theta \rightarrow \infty$ (Fig. \ref{fig:sqrt}). Accordingly, if for whatever reason we found ourselves at $\theta \gg \theta_0$, for example if the learning landscape changed rapidly upon jumping from one attractor to another, the signal in $\frac{dQ}{d \log\theta}$ could become exponentially small. Likewise, $\frac{dQ}{d \log\theta}$ diverges as $\theta \rightarrow 0$, which we felt could cause some instability. For these reasons, we feel that it is probably better to solve Eqn. \ref{best} from estimates of $S$ rather than $Q$.

\clearpage

\section{Discussion}

In this paper, we described a DSML protocol consisting of two parts: an inner loop for greedy meta-learning and an outer loop for tuning the timescales $\mu$ and $\tau$ on the fly. Our key insights -- that the intelligence is distinguishable only at an intermediate timescale, that mutation and evaluation times must be distinct, and that DSML can be boiled down to two key parameters, $\mu$ and $\tau$ -- are all contained in the inner learning cycle.

By contrast, the outer loop follows in the footsteps of many parameter optimization schemes stretching as far back as Schwefel's self-adaptation method, and it may be altered in many ways to suit the task at hand \citep{Schwefel_81}. In online learning, for example, a consecutive least-squares-type update of the form $\lambda_{i+1} = \lambda_{i} + \eta (Q_i - Q_{i-1})(\xi_i - \xi_{i-1})$ with perturbations $ \xi_i \sim \mathcal{N}(\lambda_i, \sigma^2)$ may be used \footnote{We are not entirely sure who to credit for such an optimizer. It is very similar to residual feedback \citep{Zhang_21_Zavlanos}. It is also little more than an evolution strategy (ES) or parameter-exploring policy gradient (PGPE) with a fixed variance normal distribution \citep{Sehnke_10_Schmidhuber, Wierstra_14_Schmidhuber}. These in turn are just variations on \citet{Berny_00}, which in turn is a variation on \citet{Williams_92}...}. The bipartite organization of our protocol should help facilitate this, and we hope that our results in Sec. \ref{sec:shape} will help readers to devise their own techniques which may be better than ours.

An interesting outcome of our analysis is that the learning rate $\eta$ is the only hyperparameter that requires any careful handling. The learning rate must be small enough that the parameter fluctuations are no greater than $\sigma$, yet it must be large enough that it can adapt quickly to changes in the learning landscape. Hopefully, these shifts will occur slowly enough that a wide range of $\eta$ values can be used, but if rapid changes occur, then this may turn out to be a tricky balancing act. Future refinements to the tuning algorithm may handle this problem with adaptive learning rates similar to those we have seen in stochastic gradient descent \citep{Duchi_11_Singer, Kingma_15_Ba}.

The biggest question we have left unanswered in DSML is whether superior non-greedy algorithms exist. An affirmative answer will require both theoretical and experimental results. On the theoretical side, one would need to devise a principled algorithm to obtain some big-O improvement over the greedy approach. On the experimental side, one would have to show that local optima are a serious problem for greedy DSML, and that a non-greedy algorithm can overcome this problem to such an extent as to justify its inevitable logarithmic overhead.

There is much more to discover in the budding field of DSML. We look forward to discussing how to craft the architecture and dynamics of meta-learning systems in an upcoming publication. There are also open questions as to whether modern hardware is fit for this form of machine learning, whether DSML will support or supplant gradient descent, whether DSML will come to devour as much compute as deep learning, and many more mysteries we have yet to ponder. We invite you to join us on our journey beyond the predictability horizon. These lands, it would appear, have fruit to bear, not only for ML systems, but for ML researchers as well!

\bibliography{mybib}

@inproceedings{Andrychowicz_16_De_Freitas,
    author = {Andrychowicz, Marcin and Denil, Misha and Colmenarejo, Sergio G\'{o}mez and Hoffman, Matthew W. and Pfau, David and Schaul, Tom and Shillingford, Brendan and de Freitas, Nando},
    title = {Learning to learn by gradient descent by gradient descent},
    year = {2016},
    isbn = {9781510838819},
    publisher = {Curran Associates Inc.},
    address = {Red Hook, NY, USA},
    booktitle = {Proceedings of the 30th International Conference on Neural Information Processing Systems},
    pages = {3988–3996},
    numpages = {9},
    location = {Barcelona, Spain},
    series = {NIPS'16}
}

@InProceedings{Berny_00,
    author="Berny, A.",
    editor="Schoenauer, Marc
    and Deb, Kalyanmoy
    and Rudolph, G{\"u}nther
    and Yao, Xin
    and Lutton, Evelyne
    and Merelo, Juan Julian
    and Schwefel, Hans-Paul",
    title="Selection and Reinforcement Learning for Combinatorial Optimization",
    booktitle="Parallel Problem Solving from Nature PPSN VI",
    year="2000",
    publisher="Springer Berlin Heidelberg",
    address="Berlin, Heidelberg",
    pages="601--610",
    isbn="978-3-540-45356-7"
}

@inproceedings{Brown_20_Amodei,
    author = {Brown, Tom and Mann, Benjamin and Ryder, Nick and Subbiah, Melanie and Kaplan, Jared D and Dhariwal, Prafulla and Neelakantan, Arvind and Shyam, Pranav and Sastry, Girish and Askell, Amanda and Agarwal, Sandhini and Herbert-Voss, Ariel and Krueger, Gretchen and Henighan, Tom and Child, Rewon and Ramesh, Aditya and Ziegler, Daniel and Wu, Jeffrey and Winter, Clemens and Hesse, Chris and Chen, Mark and Sigler, Eric and Litwin, Mateusz and Gray, Scott and Chess, Benjamin and Clark, Jack and Berner, Christopher and McCandlish, Sam and Radford, Alec and Sutskever, Ilya and Amodei, Dario},
    booktitle = {Advances in Neural Information Processing Systems},
    editor = {H. Larochelle and M. Ranzato and R. Hadsell and M.F. Balcan and H. Lin},
    pages = {1877--1901},
    publisher = {Curran Associates, Inc.},
    title = {Language Models are Few-Shot Learners},
    url = {https://proceedings.neurips.cc/paper_files/paper/2020/file/1457c0d6bfcb4967418bfb8ac142f64a-Paper.pdf},
    volume = {33},
    year = {2020}
}

@InProceedings{Donahue_14_Darrell,
  title = 	 {DeCAF: A Deep Convolutional Activation Feature for Generic Visual Recognition},
  author = 	 {Donahue, Jeff and Jia, Yangqing and Vinyals, Oriol and Hoffman, Judy and Zhang, Ning and Tzeng, Eric and Darrell, Trevor},
  booktitle = 	 {Proceedings of the 31st International Conference on Machine Learning},
  pages = 	 {647--655},
  year = 	 {2014},
  editor = 	 {Xing, Eric P. and Jebara, Tony},
  volume = 	 {32},
  number =       {1},
  series = 	 {Proceedings of Machine Learning Research},
  address = 	 {Bejing, China},
  month = 	 {22--24 Jun},
  publisher =    {PMLR},
  url = 	 {https://proceedings.mlr.press/v32/donahue14.html},
}

@inproceedings{Koch_15_Salakhutdinov,
  title={Siamese Neural Networks for One-Shot Image Recognition},
  author={Gregory Koch and Richard Zemel and Ruslan Salakhutdinov},
  year={2015}
}

@InProceedings{Arjovsky_16_Bengio,
  title = 	 {Unitary Evolution Recurrent Neural Networks},
  author = 	 {Arjovsky, Martin and Shah, Amar and Bengio, Yoshua},
  booktitle = 	 {Proceedings of The 33rd International Conference on Machine Learning},
  pages = 	 {1120--1128},
  year = 	 {2016},
  editor = 	 {Balcan, Maria Florina and Weinberger, Kilian Q.},
  volume = 	 {48},
  series = 	 {Proceedings of Machine Learning Research},
  address = 	 {New York, New York, USA},
  month = 	 {20--22 Jun},
  publisher =    {PMLR},
  url = 	 {https://proceedings.mlr.press/v48/arjovsky16.html}
}

@misc{Ba_16_Hinton,
      title={Layer Normalization}, 
      author={Jimmy Lei Ba and Jamie Ryan Kiros and Geoffrey E. Hinton},
      year={2016},
      eprint={1607.06450},
      archivePrefix={arXiv},
      primaryClass={stat.ML},
      url={https://arxiv.org/abs/1607.06450}, 
}

@article{Back_23_Ye,
    author = {Bäck, Thomas H. W. and Kononova, Anna V. and van Stein, Bas and Wang, Hao and Antonov, Kirill A. and Kalkreuth, Roman T. and de Nobel, Jacob and Vermetten, Diederick and de Winter, Roy and Ye, Furong},
    title = {Evolutionary Algorithms for Parameter Optimization—Thirty
                    Years Later},
    journal = {Evolutionary Computation},
    volume = {31},
    number = {2},
    pages = {81-122},
    year = {2023},
    month = {06},
    issn = {1063-6560},
    doi = {10.1162/evco_a_00325},
    url = {https://doi.org/10.1162/evco_a_00325},
    eprint = {https://direct.mit.edu/evco/article-pdf/31/2/81/2133989/evco_a_00325.pdf},
}

@article{Schmidhuber_92,
    author = {Jurgen Schmidhuber},
    title = {Steps towards `self-referential' learning: a thought experiment},
    year = {1992},
    institution = {University of Colorado Department of Computer Science},
    number = {CU-CS-627-92},
    type = {Technical Report},
}

@article{Bengio_94_Frasconi,
    author = {Bengio, Y. and Simard, P. and Frasconi, P.},
    title = {Learning long-term dependencies with gradient descent is difficult},
    year = {1994},
    issue_date = {March 1994},
    publisher = {IEEE Press},
    volume = {5},
    number = {2},
    issn = {1045-9227},
    url = {https://doi.org/10.1109/72.279181},
    doi = {10.1109/72.279181},
    month = mar,
    pages = {157–166},
    numpages = {10}
}

@article{Bengio_95_Cloutier,
    author = {Yoshua Bengio and Samy Bengio and Jocelyn Cloutier},
    title = {On the search for new learning rules for ANNs},
    journal = {Neural Processing Letters},
    volume = {2},
    pages = {26-30},
    year = {1995},
    number = {751},
    doi = {10.1007/BF02279935},
}

@article{Beyer_02_Schwefel,
    author = {Hans-Georg Beyer and Hans-Paul Schwefel},
    title = {Evolution strategies - A comprehensive introduction},
    journal = {Natural Computing},
    volume = {1},
    number = {1},
    pages = {3-52},
    year = {2002},
    doi = {10.1023/A:1015059928466},
}

@article{Chandar_19_Bengio,
  author       = {Sarath Chandar and
                  Chinnadhurai Sankar and
                  Eugene Vorontsov and
                  Samira Ebrahimi Kahou and
                  Yoshua Bengio},
  title        = {Towards Non-saturating Recurrent Units for Modelling Long-term Dependencies},
  journal      = {CoRR},
  volume       = {abs/1902.06704},
  year         = {2019},
  url          = {http://arxiv.org/abs/1902.06704},
  eprinttype   = {arXiv},
  eprint       = {1902.06704},
  bibsource    = {dblp computer science bibliography, https://dblp.org}
}

@article{Devlin_18_Toutanova,
  author       = {Jacob Devlin and
                  Ming{-}Wei Chang and
                  Kenton Lee and
                  Kristina Toutanova},
  title        = {{BERT:} Pre-training of Deep Bidirectional Transformers for Language
                  Understanding},
  journal      = {CoRR},
  volume       = {abs/1810.04805},
  year         = {2018},
  url          = {http://arxiv.org/abs/1810.04805},
  eprinttype   = {arXiv},
  eprint       = {1810.04805},
  bibsource    = {dblp computer science bibliography, https://dblp.org}
}

@article{Duchi_11_Singer,
  author  = {John Duchi and Elad Hazan and Yoram Singer},
  title   = {Adaptive Subgradient Methods for Online Learning and Stochastic Optimization},
  journal = {Journal of Machine Learning Research},
  year    = {2011},
  volume  = {12},
  number  = {61},
  pages   = {2121--2159},
  url     = {http://jmlr.org/papers/v12/duchi11a.html}
}

@article{Ermoliev_69,
  title={On the method of generalized stochastic gradients and quasi-F{\'e}jer sequences},
  author={Yu. M. Ermol’ev},
  journal={Cybernetics},
  year={1969},
  volume={5},
  pages={208-220},
  url={https://api.semanticscholar.org/CorpusID:123001116}
}

@inproceedings{Finn_17_Levine,
    author = {Finn, Chelsea and Abbeel, Pieter and Levine, Sergey},
    title = {Model-agnostic meta-learning for fast adaptation of deep networks},
    year = {2017},
    publisher = {JMLR.org},
    booktitle = {Proceedings of the 34th International Conference on Machine Learning - Volume 70},
    pages = {1126–1135},
    numpages = {10},
    location = {Sydney, NSW, Australia},
    series = {ICML'17}
}

@article{Hafner_20_Ba,
  author       = {Danijar Hafner and
                  Timothy P. Lillicrap and
                  Mohammad Norouzi and
                  Jimmy Ba},
  title        = {Mastering Atari with Discrete World Models},
  journal      = {CoRR},
  volume       = {abs/2010.02193},
  year         = {2020},
  url          = {https://arxiv.org/abs/2010.02193},
  eprinttype   = {arXiv},
  eprint       = {2010.02193},
  bibsource    = {dblp computer science bibliography, https://dblp.org}
}

@article{Hansen_95_Gawelczyk,
    author = {Nikolaus Hansen and Andreas Ostermeier and Andreas Gawelczyk},
    title = {On the adaptation of arbitrary normal mutation distributions in evolution strategies: The generating set adaptation},
    journal = {Proceedings of the Sixth International Conference on Genetic Algorithms},
    pages = {57-64},
    year = {1995},
}

@INPROCEEDINGS{Hansen_96_Ostermeier,
  author={Hansen, N. and Ostermeier, A.},
  booktitle={Proceedings of IEEE International Conference on Evolutionary Computation}, 
  title={Adapting arbitrary normal mutation distributions in evolution strategies: the covariance matrix adaptation}, 
  year={1996},
  volume={},
  number={},
  pages={312-317},
  doi={10.1109/ICEC.1996.542381}}

@ARTICLE{Hansen_01_Ostermeier,
  author={Hansen, Nikolaus and Ostermeier, Andreas},
  journal={Evolutionary Computation}, 
  title={Completely Derandomized Self-Adaptation in Evolution Strategies}, 
  year={2001},
  volume={9},
  number={2},
  pages={159-195},
  doi={10.1162/106365601750190398}}

@INPROCEEDINGS {He_16_Sun,
    author = { He, Kaiming and Zhang, Xiangyu and Ren, Shaoqing and Sun, Jian },
    booktitle = { 2016 IEEE Conference on Computer Vision and Pattern Recognition (CVPR) },
    title = {{ Deep Residual Learning for Image Recognition }},
    year = {2016},
    volume = {},
    ISSN = {1063-6919},
    pages = {770-778},
    doi = {10.1109/CVPR.2016.90},
    url = {https://doi.ieeecomputersociety.org/10.1109/CVPR.2016.90},
    publisher = {IEEE Computer Society},
    address = {Los Alamitos, CA, USA},
month =Jun
}

@article{Hestness_17_Diamos,
  author       = {Joel Hestness and
                  Sharan Narang and
                  Newsha Ardalani and
                  Gregory F. Diamos and
                  Heewoo Jun and
                  Hassan Kianinejad and
                  Md. Mostofa Ali Patwary and
                  Yang Yang and
                  Yanqi Zhou},
  title        = {Deep Learning Scaling is Predictable, Empirically},
  journal      = {CoRR},
  volume       = {abs/1712.00409},
  year         = {2017},
  url          = {http://arxiv.org/abs/1712.00409},
  eprinttype   = {arXiv},
  eprint       = {1712.00409},
  bibsource    = {dblp computer science bibliography, https://dblp.org}
}

@article{Hochreiter_97_Schmidhuber,
    author = {Hochreiter, Sepp and Schmidhuber, J\"{u}rgen},
    title = {Long Short-Term Memory},
    year = {1997},
    issue_date = {November 15, 1997},
    publisher = {MIT Press},
    address = {Cambridge, MA, USA},
    volume = {9},
    number = {8},
    issn = {0899-7667},
    url = {https://doi.org/10.1162/neco.1997.9.8.1735},
    doi = {10.1162/neco.1997.9.8.1735},
    journal = {Neural Comput.},
    month = nov,
    pages = {1735–1780},
    numpages = {46}
}

@ARTICLE{Hospedales_22_Storkey,
    author={Hospedales, Timothy and Antoniou, Antreas and Micaelli, Paul and Storkey, Amos},
    journal={ IEEE Transactions on Pattern Analysis \& Machine Intelligence },
    title={{ Meta-Learning in Neural Networks: A Survey }},
    year={2022},
    volume={44},
    number={09},
    ISSN={1939-3539},
    pages={5149-5169},
    doi={10.1109/TPAMI.2021.3079209},
    url = {https://doi.ieeecomputersociety.org/10.1109/TPAMI.2021.3079209},
    publisher={IEEE Computer Society},
    address={Los Alamitos, CA, USA},
    month=sep
}

@article{Howard_18_Ruder,
  author       = {Jeremy Howard and
                  Sebastian Ruder},
  title        = {Fine-tuned Language Models for Text Classification},
  journal      = {CoRR},
  volume       = {abs/1801.06146},
  year         = {2018},
  url          = {http://arxiv.org/abs/1801.06146},
  eprinttype   = {arXiv},
  eprint       = {1801.06146},
  bibsource    = {dblp computer science bibliography, https://dblp.org}
}

@article{Hubert_26_Silver,
    author = {Hubert, Thomas and Mehta, Rishi and Sartran, Laurent and Horváth, Miklós Z. and Žužić, Goran and Wieser, Eric and Huang, Aja and Schrittwieser, Julian and Schroecker, Yannick and Masoom, Hussain and Bertolli, Ottavia and Zahavy, Tom and Mandhane, Amol and Yung, Jessica and Beloshapka, Iuliya and Ibarz, Borja and Veeriah, Vivek and Yu, Lei and Nash, Oliver and Lezeau, Paul and Mercuri, Salvatore and Sönne, Calle and Mehta, Bhavik and Davies, Alex and Zheng, Daniel and Pedregosa, Fabian and Li, Yin and von Glehn, Ingrid and Rowland, Mark and Albanie, Samuel and Velingker, Ameya and Schmitt, Simon and Lockhart, Edward and Hughes, Edward and Michalewski, Henryk and Sonnerat, Nicolas and Hassabis, Demis and Kohli, Pushmeet and Silver, David},
    title = {Olympiad-level formal mathematical reasoning with reinforcement learning},
    journal = {Nature},
    volume = {651},
    pages = {607--613},
    year = {2025},
    doi = {10.1038/s41586-025-09833-y},
    url = {https://doi.org/10.1038/s41586-025-09833-y}
}

@article{Huisman_21_Plaat,
    author = {Huisman, Mike and van Rijn, Jan N. and Plaat, Aske},
    title = {A survey of deep meta-learning},
    year = {2021},
    issue_date = {Aug 2021},
    publisher = {Kluwer Academic Publishers},
    address = {USA},
    volume = {54},
    number = {6},
    issn = {0269-2821},
    url = {https://doi.org/10.1007/s10462-021-10004-4},
    doi = {10.1007/s10462-021-10004-4},
    journal = {Artif. Intell. Rev.},
    month = aug,
    pages = {4483–4541},
    numpages = {59},
}

@article{Kaplan_20_Amodei,
  author       = {Jared Kaplan and
                  Sam McCandlish and
                  Tom Henighan and
                  Tom B. Brown and
                  Benjamin Chess and
                  Rewon Child and
                  Scott Gray and
                  Alec Radford and
                  Jeffrey Wu and
                  Dario Amodei},
  title        = {Scaling Laws for Neural Language Models},
  journal      = {CoRR},
  volume       = {abs/2001.08361},
  year         = {2020},
  url          = {https://arxiv.org/abs/2001.08361},
  eprinttype   = {arXiv},
  eprint       = {2001.08361},
  bibsource    = {dblp computer science bibliography, https://dblp.org}
}

@article{Kingma_15_Ba,
  title={Adam: A Method for Stochastic Optimization},
  author={Kingma, Diederik P. and Ba, Jimmy},
  journal={International Conference on Learning Representations},
  year={2015}
}

@book{Kushner_78_Clark,
  title={Stochastic Approximation Methods for Constrained and Unconstrained Systems},
  author={Kushner, Harold J. and Clark, David S.},
  year={1978},
  publisher={Springer-Verlag New York, Inc.},
  address={New York, NY, USA}
}

@article{Le_15_Hinton,
  author       = {Quoc V. Le and
                  Navdeep Jaitly and
                  Geoffrey E. Hinton},
  title        = {A Simple Way to Initialize Recurrent Networks of Rectified Linear Units},
  journal      = {CoRR},
  volume       = {abs/1504.00941},
  year         = {2015},
  url          = {http://arxiv.org/abs/1504.00941},
  eprinttype   = {arXiv},
  eprint       = {1504.00941},
  bibsource    = {dblp computer science bibliography, https://dblp.org}
}

@article{Li_17_Malik,
  author       = {Ke Li and
                  Jitendra Malik},
  title        = {Learning to Optimize Neural Nets},
  journal      = {CoRR},
  volume       = {abs/1703.00441},
  year         = {2017},
  url          = {http://arxiv.org/abs/1703.00441},
  eprinttype   = {arXiv},
  eprint       = {1703.00441},
  bibsource    = {dblp computer science bibliography, https://dblp.org}
}

@InProceedings{Maheswaranathan_19_Sohl_Dickstein,
  title = 	 {Guided evolutionary strategies: augmenting random search with surrogate gradients},
  author =       {Maheswaranathan, Niru and Metz, Luke and Tucker, George and Choi, Dami and Sohl-Dickstein, Jascha},
  booktitle = 	 {Proceedings of the 36th International Conference on Machine Learning},
  pages = 	 {4264--4273},
  year = 	 {2019},
  editor = 	 {Chaudhuri, Kamalika and Salakhutdinov, Ruslan},
  volume = 	 {97},
  series = 	 {Proceedings of Machine Learning Research},
  month = 	 {09--15 Jun},
  publisher =    {PMLR},
  url = 	 {https://proceedings.mlr.press/v97/maheswaranathan19a.html},
}

@article{Merity_17_Socher,
  author       = {Stephen Merity and
                  Nitish Shirish Keskar and
                  Richard Socher},
  title        = {Regularizing and Optimizing {LSTM} Language Models},
  journal      = {CoRR},
  volume       = {abs/1708.02182},
  year         = {2017},
  url          = {http://arxiv.org/abs/1708.02182},
  eprinttype   = {arXiv},
  eprint       = {1708.02182},
  bibsource    = {dblp computer science bibliography, https://dblp.org}
}

@inproceedings{Pascanu_13_Bengio,
    author = {Pascanu, Razvan and Mikolov, Tomas and Bengio, Yoshua},
    title = {On the difficulty of training recurrent neural networks},
    year = {2013},
    publisher = {JMLR.org},
    booktitle = {Proceedings of the 30th International Conference on International Conference on Machine Learning - Volume 28},
    pages = {III–1310–III–1318},
    location = {Atlanta, GA, USA},
    series = {ICML'13}
}

@article{Radford_19_Sutskever,
  author       = {Alec Radford and
                  Jeffrey Wu and
                  Rewon Child and
                  David Luan and
                  Dario Amodei and
                  Ilya Sutskever},
  title        = {Language Models are Unsupervised Multitask Learners},
  year = {2019}
}

@book{Rechenberg_73,
  author    = {Ingo Rechenberg},
  title     = {Evolutionsstrategie: Optimierung technischer Systeme nach
               Prinzipien der biologischen Evolution},
  year      = {1973},
  publisher = {Frommann-Holzboog},
  address   = {Stuttgart}
}

@InProceedings{Ros_08_Hansen,
  author="Ros, Raymond
  and Hansen, Nikolaus",
  editor="Rudolph, G{\"u}nter
  and Jansen, Thomas
  and Beume, Nicola
  and Lucas, Simon
  and Poloni, Carlo",
  title="A Simple Modification in CMA-ES Achieving Linear Time and Space Complexity",
  booktitle="Parallel Problem Solving from Nature -- PPSN X",
  year="2008",
  publisher="Springer Berlin Heidelberg",
  address="Berlin, Heidelberg",
  pages="296--305",
  isbn="978-3-540-87700-4"
}

@INPROCEEDINGS{Runarsson_00_Jonsson,
  author={Runarsson, T.P. and Jonsson, M.T.},
  booktitle={2000 IEEE Symposium on Combinations of Evolutionary Computation and Neural Networks. Proceedings of the First IEEE Symposium on Combinations of Evolutionary Computation and Neural Networks (Cat. No.00}, 
  title={Evolution and design of distributed learning rules}, 
  year={2000},
  volume={},
  number={},
  pages={59-63},
  doi={10.1109/ECNN.2000.886220}}

@inproceedings{Schaeffer_23_Koyejo,
  title={Are Emergent Abilities of Large Language Models a Mirage?},
  author={Rylan Schaeffer and Brando Miranda and Sanmi Koyejo},
  booktitle={Thirty-seventh Conference on Neural Information Processing Systems},
  year={2023},
  url={https://openreview.net/forum?id=ITw9edRDlD}
}

@INPROCEEDINGS{Schmidhuber_93_meta,
  author={Schmidhuber, J.},
  booktitle={IEEE International Conference on Neural Networks}, 
  title={A neural network that embeds its own meta-levels}, 
  year={1993},
  volume={},
  number={},
  pages={407-412 vol.1},
  doi={10.1109/ICNN.1993.298591}
}

@article{Schmidhuber_99_Wiering,
    author = {Schmidhuber, J and Zhao, Jieyu and Wiering, Marco},
    year = {1999},
    month = {02},
    title = {Simple Principles Of Metalearning}
}

@ARTICLE{Schrittwieser_20_Silver,
    author={Schrittwieser, Julian and Antonoglou, Ioannis and Hubert, Thomas and Simonyan, Karen and Sifre, Laurent and Schmitt, Simon and Guez, Arthur and Lockhart, Edward and Hassabis, Demis and Graepel, Thore and Lillicrap, Timothy and Silver, David},
    journal={Nature}, 
    title={Mastering Atari, Go, chess and shogi by planning with a learned model}, 
    year={2020},
    volume={588},
    pages={604-609},
    doi={10.1038/s41586-020-03051-4},
}

@book{Schwefel_81,
  author    = {Hans-Paul Schwefel},
  title     = {Numerical Optimization of Computer Models},
  publisher = {Wiley},
  address   = {Chichester},
  year      = {1981}
}

@article{Sehnke_10_Schmidhuber,
    title = {Parameter-exploring policy gradients},
    journal = {Neural Networks},
    volume = {23},
    number = {4},
    pages = {551-559},
    year = {2010},
    note = {The 18th International Conference on Artificial Neural Networks, ICANN 2008},
    issn = {0893-6080},
    doi = {https://doi.org/10.1016/j.neunet.2009.12.004},
    url = {https://www.sciencedirect.com/science/article/pii/S0893608009003220},
    author = {Frank Sehnke and Christian Osendorfer and Thomas Rückstieß and Alex Graves and Jan Peters and Jürgen Schmidhuber}
}

@ARTICLE{Silver_17_Hassabis,
    author={Silver, David and Schrittwieser, Julian and Simonyan, Karen and Antonoglou, Ioannis and Huang, Aja and Guez, Arthur and Hubert, Thomas and Baker, Lucas and Lai, Matthew and Bolton, Adrian and Chen, Yutian and Lillicrap, Timothy and Hui, Fan and Sifre, Laurent and van den Driessche, George and Graepel, Thore and Hassabis, Demis},
    journal={Nature}, 
    title={Mastering the game of Go without human knowledge}, 
    year={2017},
    volume={550},
    pages={354-359},
    doi={10.1038/nature24270},
}

@article{Srivastava_15_Schmidhuber,
  author       = {Rupesh Kumar Srivastava and
                  Klaus Greff and
                  J{\"{u}}rgen Schmidhuber},
  title        = {Highway Networks},
  journal      = {CoRR},
  volume       = {abs/1505.00387},
  year         = {2015},
  url          = {http://arxiv.org/abs/1505.00387},
  eprinttype   = {arXiv},
  eprint       = {1505.00387},
  bibsource    = {dblp computer science bibliography, https://dblp.org}
}

@misc{Sutton_19,
  author = {Sutton, Richard S.},
  title = {The Bitter Lesson},
  year = {2019},
  month = mar,
  day = {13},
  howpublished = {\url{http://www.incompleteideas.net/IncIdeas/BitterLesson.html}},
  note = {Accessed: 2026-06-18}
}

@Inbook{Thrun_98_Pratt,
    author="Thrun, Sebastian
    and Pratt, Lorien",
    editor="Thrun, Sebastian
    and Pratt, Lorien",
    title="Learning to Learn: Introduction and Overview",
    bookTitle="Learning to Learn",
    year="1998",
    publisher="Springer US",
    address="Boston, MA",
    pages="3--17",
    isbn="978-1-4615-5529-2",
    doi="10.1007/978-1-4615-5529-2_1",
    url="https://doi.org/10.1007/978-1-4615-5529-2_1"
}

@article{Veit_16_Belongie,
  author       = {Andreas Veit and
                  Michael J. Wilber and
                  Serge J. Belongie},
  title        = {Residual Networks are Exponential Ensembles of Relatively Shallow
                  Networks},
  journal      = {CoRR},
  volume       = {abs/1605.06431},
  year         = {2016},
  url          = {http://arxiv.org/abs/1605.06431},
  eprinttype   = {arXiv},
  eprint       = {1605.06431},
  bibsource    = {dblp computer science bibliography, https://dblp.org}
}

@article{Vettoruzzo_24_Thorsteinn,
    author = {Vettoruzzo, Anna and Bouguelia, Mohamed-Rafik and Vanschoren, Joaquin and R\"{o}gnvaldsson, Thorsteinn and Santosh, KC},
    title = {Advances and Challenges in Meta-Learning: A Technical Review},
    year = {2024},
    issue_date = {July 2024},
    publisher = {IEEE Computer Society},
    address = {USA},
    volume = {46},
    number = {7},
    issn = {0162-8828},
    url = {https://doi.org/10.1109/TPAMI.2024.3357847},
    doi = {10.1109/TPAMI.2024.3357847},
    journal = {IEEE Trans. Pattern Anal. Mach. Intell.},
    month = jul,
    pages = {4763–4779},
    numpages = {17}
}

@misc{Wang_16_Freitas,
      title={Bayesian Optimization in a Billion Dimensions via Random Embeddings}, 
      author={Ziyu Wang and Frank Hutter and Masrour Zoghi and David Matheson and Nando de Freitas},
      year={2016},
      eprint={1301.1942},
      archivePrefix={arXiv},
      primaryClass={stat.ML},
      url={https://arxiv.org/abs/1301.1942}, 
}

@misc{Wei_22_Fedus,
  title={Emergent Abilities of Large Language Models},
  author={Jason Wei and Yi Tay and Rishi Bommasani and Colin Raffel and Barret Zoph and Sebastian Borgeaud and Dani Yogatama and Maarten Bosma and Denny Zhou and Donald Metzler and Ed H. Chi and Tatsunori Hashimoto and Oriol Vinyals and Percy Liang and Jeff Dean and William Fedus},
  journal={Transactions on Machine Learning Research},
  issn={2835-8856},
  year={2022},
  url={https://openreview.net/forum?id=yzkSU5zdwD},
  note={Survey Certification}
}

@article{Wierstra_14_Schmidhuber,
  author  = {Daan Wierstra and Tom Schaul and Tobias Glasmachers and Yi Sun and Jan Peters and Jurgen Schmidhuber},
  title   = {Natural Evolution Strategies},
  journal = {Journal of Machine Learning Research},
  year    = {2014},
  volume  = {15},
  number  = {27},
  pages   = {949--980},
  url     = {http://jmlr.org/papers/v15/wierstra14a.html}
}

@article{Williams_92,
  author  = {R.J. Williams},
  title   = {Simple statistical gradient-following algorithms for connectionist reinforcement learning},
  journal = {Mach Learn},
  year    = {1992},
  volume  = {8},
  pages   = {229–256},
  doi     = {10.1007/BF00992696}
}

@article{Yosinski_14_Lipson,
  author       = {Jason Yosinski and
                  Jeff Clune and
                  Yoshua Bengio and
                  Hod Lipson},
  title        = {How transferable are features in deep neural networks?},
  journal      = {CoRR},
  volume       = {abs/1411.1792},
  year         = {2014},
  url          = {http://arxiv.org/abs/1411.1792},
  eprinttype   = {arXiv},
  eprint       = {1411.1792},
  bibsource    = {dblp computer science bibliography, https://dblp.org}
}

@article{Zaremba_14_Vinyals,
  author       = {Wojciech Zaremba and
                  Ilya Sutskever and
                  Oriol Vinyals},
  title        = {Recurrent Neural Network Regularization},
  journal      = {CoRR},
  volume       = {abs/1409.2329},
  year         = {2014},
  url          = {http://arxiv.org/abs/1409.2329},
  eprinttype   = {arXiv},
  eprint       = {1409.2329},
  bibsource    = {dblp computer science bibliography, https://dblp.org}
}

@misc{Zhang_21_Zavlanos,
      title={A New One-Point Residual-Feedback Oracle For Black-Box Learning and Control}, 
      author={Yan Zhang and Yi Zhou and Kaiyi Ji and Michael M. Zavlanos},
      year={2021},
      eprint={2006.10820},
      archivePrefix={arXiv},
      primaryClass={math.OC},
      url={https://arxiv.org/abs/2006.10820}, 
}
\bibliographystyle{tmlr-style-file-main/tmlr}

\end{document}